\documentclass[11pt,a4paper]{article}
\usepackage[hyperref]{eacl2021}
\usepackage{times}
\usepackage{latexsym}

\usepackage{graphicx}
\usepackage{microtype}

\aclfinalcopy 

\title{NaijaNER : Comprehensive Named Entity Recognition for 5 Nigerian Languages.}
\author{\\\\ \textbf{Wuraola Fisayo Oyewusi,  Olubayo Adekanmbi, Ifeoma Okoh, Vitus Onuigwe,}\\ \textbf{Mary Idera Salami, Opeyemi Osakuade, Sharon Ibejih, Usman Abdullahi Musa,}\\\\
Data Science Nigeria, Lagos Nigeria\\ \{wuraola, olubayo, ifeoma, sharon, mary, abdullahi, osakuade, vitus\}@datasciencenigeria.ai}

\date{}

\begin{document}
\maketitle
\begin{abstract}

Most of the common applications of Named Entity Recognition (NER) is on English and other highly available languages. In this work, we present our findings on Named Entity Recognition for 5 Nigerian Languages (Nigerian English, Nigerian Pidgin English, Igbo, Yoruba and Hausa). These languages are  considered low-resourced, and very little openly available Natural Language Processing work has been done in most of them. In this work, individual NER models were trained and metrics recorded for each of the languages. We also worked on a combined model that can handle Named Entity Recognition (NER) for any of the five languages. The combined model works  well for Named Entity Recognition(NER) on each of the languages and with better performance compared to individual NER models trained specifically on annotated data for the specific language. The aim of this work is to share our learning on how information extraction using Named Entity Recognition can be optimized for the listed Nigerian Languages for inclusion, ease of deployment in production and reusability of models. Models developed during this project are available on \href{https://git.io/JY0kk}{GitHub} and an interactive web app \href{https://nigner.herokuapp.com/}{here}.

\end{abstract} 

\section{Introduction}

Named entity recognition(NER) is sometimes referred to as entity chunking, extraction, or identification. It is a subtask of information extraction \citep{ff(2007)} in Natural Language Processing. Named Entity recognition involves processing a text and identifying certain occurrences of words or expressions as belonging to particular categories of Named Entities (NE) \citep{r3(1999)}. Named entity recognition (NER) is commonly dealt with as a sequence labelling job and has been one of the most successfully tackled NLP tasks \citep{r2(2021)}.

The phrase “Named Entity” was coined in 1996 at the 6th Message Understanding Conference (MUC) when the extraction of information from unstructured text became an important problem in the linguistic domain \citep{ff(2007)}

The concept of Named entity recognition transcends language and the principle is the same. Natural Language Processing enables improved communication between human and machines. Some of the most important use cases for Named Entity Recognition involve large text data. These use cases range from recognizing Named Entities for document classification, entity linking, social media analysis and methods for training NER models. For example \citep{Clark(2011)} in their work, “Named Entity Recognition in Tweets'' address the approaches and challenges related to the style of unintentional language being used in social media, and how the limitation of standard NLP tool degradation on tweets and their custom T-NER model gave better performance. This makes a case for context-specific training of NER models.  \citep{Shinyama&Sekine(2004)}  in their work “Named Entity Discovery Using Comparable News Articles'' studied the pattern of NER in news articles and the way common nouns are used. The techniques used involve classifying based on the time series distribution of the entities.

\subsection{Named Entity Recognition for Low Resource Languages}
There are about 7,000+ languages spoken across the world, but of these, only a small fraction are considered high-resource languages. Languages are called low-resource when lacking large monolingual or parallel corpora and/or manually crafted linguistic resources sufficient for building statistical NLP applications \citep{sciforce_2019}. 
Many Nigerian languages are considered a low resource, including 4 of the 5 languages listed in this project. Nigerian English is not considered low resource because it is standard English but included in this project to allow for context-specific NER for entities peculiar to the Nigerian scenario.

Cross-lingual transfer learning is a technique which can compensate for the dearth of data, by transferring knowledge from high to low-resource languages. This has typically taken the form of annotation projection over parallel corpora or other multilingual resources or making use of transferable representations, such as phonetic transcriptions, closely related languages or bilingual dictionaries \citep{Rahimi_et_al(2019)}.

According to \citep{Hu_et_al(2020)}, there is still a sizable gap in the performance of cross-lingually transferred models, particularly on syntactic and sentence retrieval tasks when compared to the English Language. This makes a case for ground-up manually annotated and context-specific data for NLP tasks by native speakers. This work aims to explore training NER models for Nigerian English, Nigerian Pidgin English, Igbo, Hausa and Yoruba Languages.





\section{Methodology}

\begin{figure}[hbt!]
    \centering
    \includegraphics[width=8cm]{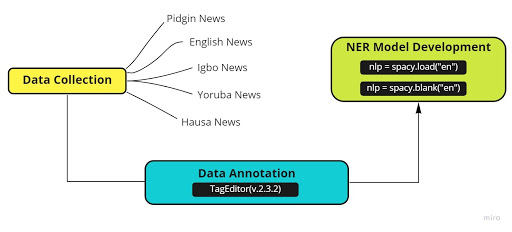}
    \caption{\label{Methodology} Methodology for training NER Models for Nigerian Languages}
\end{figure}

Figure\ref{Methodology} shows a simple flow for training Named Entity Recognition for the Nigerian Languages explored in this work.

\subsection{Data Collection}
50 articles were selected from a collection of online news articles for the 5 languages. The articles collected were in two major categories,  General and Sports News with a split ratio of 4:1

\subsection{Data Annotation}
TagEditor software (v.2.3.2) an open-source annotation tool compatible with spaCy, was used by data annotators who are expert speakers of each language to label the named entities in their respective languages. These annotation gazettes were labelled based on the \href{https://catalog.ldc.upenn.edu/LDC2013T19}{OntoNotes} Entities and description as shown in Table \ref{font-table}. The annotated data for each language was then exported and formatted to the spaCy’s {\small\verb|gold.doc_to_json|}. This is the acceptable data format for {\small\verb|spacy train|}.

\begin{table}[hbt!]
\centering
\scalebox{0.54}{\begin{tabular}{|p{4cm}|p{6cm}|}
\hline \textbf{OntoNotes Named Entities} & \textbf{Description} \\ \hline
PERSON & People, including fictional. \\
NORP & Nationalities or religious or political groups. \\
FAC & Buildings, airports, highways, bridges, etc.\\
ORG & Companies, agencies, institutions, etc. \\
GPE & Countries, cities, states. \\
LOC & Names of geographical locations other than GPEs. \\
PRODUCT & Objects, vehicles, foods, etc. (Not services.)\\
EVENT & Named hurricanes, battles, wars, sports events, etc. \\
WORK\_OF\_ART & Titles of books, songs, etc. \\
LAW & Named documents made into laws. \\
LANGUAGE & Any named language. \\
DATE & Absolute or relative dates or periods. \\
TIME & Times smaller than a day. \\
PERCENT & Percentage, including \%. \\
MONEY & Monetary values, including unit. \\
QUANTITY & Measurements, as of weight or distance. \\
ORDINAL & “first”, “second”, etc. \\
CARDINAL & Numerals that do not fall under another type. \\
\hline
\end{tabular}}
\caption{\label{font-table} spaCy OntoNotes Named Entities}
\end{table}

\subsection{NER Model Development}
The approach for developing the models used in this research paper was based on spaCy. spaCy \footnote{\url{https://spacy.io/models}} is an open-source software library for advanced natural language processing, written in the programming languages Python and Cython \citep{neumann2019scispacy(2019)}. 

spaCy’s Named Entity Recognition system features a sophisticated word embedding strategy using subword features and Bloom embedding strategy, a deep learning Convolutional Neural Network with residual connections and a novel transition-based approach (which involves layer normalization and maxout non-linearity) to Named Entity parsing, giving much better efficiency than the standard BiLSTM solution \footnote{\url{https://www.youtube.com/watch?v=sqDHBH9IjRU&t=2793s}}.

spaCy takes in the annotated training dataset, which is the collection of documents labelled into existing entities. The algorithm uses machine learning to study patterns from the given set of training data.


For this project, the annotated data was split into 80\% training and 20\% evaluation sets.
Table \ref{citation-guide} represents number of sentences annotated (spaCy docs), which is different for every language. They were used while training the Named Entity Recognition(NER) models on the custom annotated data for each of the languages. Igbo, Yoruba and Hausa NER models were trained on a blank spaCy model while the base NER models for Nigerian English and Pidgin ran on the spaCy “en” pretrained model. The latter was adopted since either of the language vocabularies had similarity with spaCy’s English NER Model. The metrics used in model assessment are Precision, Recall and F1-Score

\begin{table}[ht]
\centering
\scalebox{0.7}{\begin{tabular}{ |p{3cm}|p{3cm}|p{3cm}| }
\hline
\textbf{NER Model} & \textbf{No. of Training sentences} & \textbf{No. of Evaluation sentences}\\
\hline
PidginNER & 1148 & 105 \\
NigerianEnglishNER & 312 & 95 \\
IgboNER & 502 & 196  \\
YorubaNER & 549 & 197 \\
HausaNER & 2658 & 1046 \\
NaijaNER & 5697 & 1135 \\
\hline
\end{tabular}}
\caption{\label{citation-guide}
NER Model Training Parameters
}
\end{table}


\section{Results}

Tables \ref{allner-metrics} ---\ref{naijaner-metrics} show the detailed number of entities, entity classification F1 score, Precision and Recall scores for each entity for individual language models and the combined NER model. Each table is named with the language then a suffix---NER.

\begin{table}[ht]
\centering
\textbf{Results for 50 annotated articles (80\% General News, 20\% Sports News)} \\
\scalebox{0.7}{\begin{tabular}{ |p{3cm}|p{1.5cm}|p{1.5cm}|p{1.5cm}|p{1.5cm}|}
\hline
\textbf{NER Model} & \textbf{Total Number of words} & \textbf{F1 Score} & \textbf{Precision} & \textbf{Recall}\\
\hline
PidginNER & 24716 & 67.13 & 70.29 & 64.24 \\
NigerianEnglishNER & 7246 & 54.49 & 55.48 & 53.53 \\
IgboNER & 18577 & 35.48 & 47.18 & 28.43 \\
YorubaNER & 17169 & 45.57 & 53.64 & 39.61 \\
HausaNER & 20265 & 44.84 & 52.59 & 39.08 \\
NaijaNER & 87710 & 53.75 & 62.43 & 47.19 \\
\hline
\end{tabular}}
\caption{\label{allner-metrics}
Precision, Recall and F1 Scores for each model
}
\end{table}

\begin{table}[hbt!]
\center
\scalebox{0.6}{\begin{tabular}{ |p{3cm}|p{1.4cm}|p{1.4cm}|p{1.4cm}|p{1.4cm}|}
\hline
\textbf{Entity Type} & \textbf{No. of examples per entity} & \textbf{F1 Score} & \textbf{Precision} & \textbf{Recall}\\
\hline
CARDINAL & 9 & 0.00 & 0.00 & 0.00 \\
DATE & $>$50 & 72.43 & 70.53 & 74.44 \\
EVENT & 49 & 8.33 & 50.00 & 4.55 \\
FAC & 25 & 0.00 & 0.00 & 0.00 \\
GPE & $>$50 & 55.17 & 52.59 & 39.08 \\
LANGUAGE & 6 & 53.75 & 62.43 & 47.19 \\
LAW & -- & -- & -- & -- \\
LOC & 25 & 0.00 & 0.00 & 0.00 \\
MONEY & 25 & 33.33 & 33.33 & 33.33 \\
NORP & $>$50 & 40.45 & 50.00 & 33.96 \\
ORDINAL & 34 & 25.81 & 30.77 & 22.22 \\
ORG & $>$50 & 27.96 & 24.53 & 32.50 \\
PERCENT & 4 & 0.00 & 0.00 & 0.00 \\
PERSON & $>$50 & 53.58 & 58.90 & 49.14 \\
PRODUCT & 31 & 0.00 & 0.00 & 0.00 \\
QUANTITY & $>$50 & 20.69 & 25.00 & 17.65\\
TIME & 7 & 0.00 & 0.00 & 0.00 \\
WORK\_OF\_ART & 7 & 0.00 & 0.00 & 0.00 \\
\hline
\end{tabular}}
\caption{\label{yorubaner-metrics}
Entity Evaluation Scores of the YorubaNER Model
}
\end{table}

\begin{table}[hbt!]
\center
\scalebox{0.6}{\begin{tabular}{ |p{3cm}|p{1.4cm}|p{1.4cm}|p{1.4cm}|p{1.4cm}|}
\hline
\textbf{Entity Type} & \textbf{No. of examples per entity} & \textbf{F1 Score} & \textbf{Precision} & \textbf{Recall}\\
\hline
CARDINAL & $>$50 & 38.36 & 37.84 & 38.89 \\
DATE & $>$50 & 39.44 & 40.00 & 38.89 \\
EVENT & 28 & 42.11 & 66.67 & 30.77 \\
FAC & 1 & 0.00 & 0.00 & 0.00 \\
GPE & $>$50 & 56.67 & 56.67 & 56.67 \\
LANGUAGE & 44 & 72.73 & 100.00 & 57.14 \\
LAW & $>$50 & 0.00 & 0.00 & 0.00 \\
LOC & 7 & 0.00 & 0.00 & 0.00 \\
MONEY & 15 & 33.33 & 33.33 & 33.33 \\
NORP & 34 & 44.45 & 60.00 & 35.29 \\
ORDINAL & 9 & 0.00 & 0.00 & 0.00 \\
ORG & $>$50 & 61.14 & 54.13 & 70.24 \\
PERCENT & 1 & 0.00 & 0.00 & 0.00 \\
PERSON & $>$50 & 42.07 & 59.09 & 32.66 \\
PRODUCT & 14 & 0.00 & 0.00 & 0.00 \\
QUANTITY & 1 & 0.00 & 0.00 & 0.00 \\
TIME & 26 & 0.00 & 0.00 & 0.00 \\
WORK\_OF\_ART & $>$50 & 23.88 & 40.00 & 17.02 \\
\hline
\end{tabular}}
\caption{\label{hausaner-metrics}
Entity Evaluation Scores of the HausaNER Model
}
\end{table}

\begin{table}[hbt!]
\center
\scalebox{0.6}{\begin{tabular}{ |p{3cm}|p{1.4cm}|p{1.4cm}|p{1.4cm}|p{1.4cm}|}
\hline
\textbf{Entity Type} & \textbf{No. of examples per entity} & \textbf{F1 Score} & \textbf{Precision} & \textbf{Recall}\\
\hline
CARDINAL & $>$50 & 53.85 & 48.28 & 60.87 \\
DATE & $>$50 & 46.60 & 50.00 & 43.64 \\
EVENT & 11 & 36.36 & 66.67 & 25.00 \\
FAC & 22 & 11.76 & 9.09 & 16.67 \\
GPE & $>$50 & 30.08 & 45.45 & 22.47 \\
LANGUAGE & 9 & 0.00 & 0.00 & 0.00 \\
LAW & 1 & 0.00 & 0.00 & 0.00 \\
LOC & 7 & 0.00 & 0.00 & 0.00 \\
MONEY & 26 & 35.30 & 30.00 & 42.86 \\
NORP & $>$50 & 0.00 & 0.00 & 0.00 \\
ORDINAL & 17 & 10.53 & 12.50 & 9.09 \\
ORG & $>$50 & 50.26 & 66.67 & 40.34 \\
PERCENT & 11 & 0.00 & 0.00 & 0.00 \\
PERSON & $>$50 & 37.88 & 39.06 & 36.76 \\
PRODUCT & $>$50 & 24.58 & 62.86 & 15.28 \\
QUANTITY & 27 & 0.00 & 0.00 & 0.00 \\
TIME & 35 & 87.50 & 82.35 & 93.33 \\
WORK\_OF\_ART & 7 & 0.00 & 0.00 & 0.00 \\
\hline
\end{tabular}}
\caption{\label{igboner-metrics}
Entity Evaluation Scores of the IgboNER Model
}
\end{table}

\begin{table}[hbt!]
\center
\scalebox{0.6}{\begin{tabular}{ |p{3cm}|p{1.5cm}|p{1.5cm}|p{1.5cm}|p{1.5cm}|}
\hline
\textbf{Entity Type} & \textbf{No. of examples per entity} & \textbf{F1 Score} & \textbf{Precision} & \textbf{Recall}\\
\hline
CARDINAL & $>$50 & 82.60 & 80.85 & 84.44 \\
DATE & $>$50 & 68.49 & 65.79 & 71.43 \\
EVENT & $>$50 & 71.43 & 71.43 & 71.43 \\
FAC & $>$50 & 0.00 & 0.00 & 0.00 \\
GPE & $>$50 & 53.33 & 80.00 & 40.00 \\
LANGUAGE & 5 & 0.00 & 0.00 & 0.00 \\
LAW & 44 & 0.00 & 0.00 & 0.00 \\
LOC & 55 & 0.00 & 0.00 & 0.00 \\
MONEY & $>$50 & 0.00 & 0.00 & 0.00 \\
NORP & $>$50 & 22.22 & 12.5 & 100.00 \\
ORDINAL & $>$50 & 100.00 & 100.00 & 100.00 \\
ORG & $>$50 & 65.15 & 70.00 & 60.87 \\
PERCENT & 8 & 0.00 & 0.00 & 0.00 \\
PERSON & $>$50 & 67.94 & 71.72 & 64.55 \\
PRODUCT & 14 & 0.00 & 0.00 & 0.00 \\
QUANTITY & 45 & 50.00 & 33.33 & 100.00 \\
TIME & 21 & 66.67 & 100.00 & 50.00 \\
WORK\_OF\_ART & $>$50 & 0.00 & 0.00 & 0.00 \\
\hline
\end{tabular}}
\caption{\label{pidginer-metrics}
Entity Evaluation Scores of the PidginNER Model
}
\end{table}

\begin{table}[hbt!]
\center
\scalebox{0.6}{\begin{tabular}{ |p{3cm}|p{1.5cm}|p{1.5cm}|p{1.5cm}|p{1.5cm}|}
\hline
\textbf{Entity Type} & \textbf{No. of examples per entity} & \textbf{F1 Score} & \textbf{Precision} & \textbf{Recall}\\
\hline
CARDINAL & $>$50 & 75.00 & 80.77 & 70.00 \\
DATE & $>$50 & 59.26 & 57.14 & 61.54 \\
EVENT & 34 & 42.11 & 57.14 & 33.33 \\
FAC & 7 & 0.00 & 0.00 & 0.00 \\
GPE & $>$50 & 69.84 & 68.75 & 70.97 \\
LANGUAGE & 2 & 0.00 & 0.00 & 0.00 \\
LAW & -- & -- & -- & -- \\
LOC & 5 & 0.00 & 0.00 & 0.00 \\
MONEY & 5 & 0.00 & 0.00 & 0.00 \\
NORP & $>$50 & 63.64 & 53.85 & 77.78 \\
ORDINAL & 13 & 66.67 & 66.67 & 66.67 \\
ORG & $>$50 & 29.10 & 25.81 & 33.33 \\
PERCENT & 8 & 0.00 & 0.00 & 0.00 \\
PERSON & $>$50 & 43.14 & 50.00 & 37.93 \\
PRODUCT & 1 & 0.00 & 0.00 & 0.00 \\
QUANTITY & $>$50 & 50.00 & 33.33 & 100.00 \\
TIME & 4 & 0.00 & 0.00 & 0.00 \\
WORK\_OF\_ART & 3 & 0.00 & 0.00 & 0.00 \\
\hline
\end{tabular}}
\caption{\label{Ngenglishner-metrics}
Entity Evaluation Scores of the NigerianEnglishNER Model
}
\end{table}

\begin{table}[hbt!]
\center
\scalebox{0.6}{\begin{tabular}{ |p{3cm}|p{1.5cm}|p{1.5cm}|p{1.5cm}|p{1.5cm}|}
\hline
\textbf{Entity Type} & \textbf{No. of examples per entity} & \textbf{F1 Score} & \textbf{Precision} & \textbf{Recall}\\
\hline
CARDINAL & $>$50 & 61.66 & 57.35 & 66.67 \\
DATE & $>$50 & 63.47 & 71.26 & 57.21 \\
EVENT & $>$50 & 38.41 & 65.91 & 27.10 \\
FAC & $>$50 & 0.00 & 0.00 & 0.00 \\
GPE & $>$50 & 60.10 & 66.48 & 54.84 \\
LANGUAGE & $>$50 & 85.71 & 85.71 & 85.71 \\
LAW & 28 & 0.00 & 0.00 & 0.00 \\
LOC & $>$50 & 8.70 & 25.0 & 5.26 \\
MONEY & $>$50 & 0.00 & 0.00 & 0.00 \\
NORP & $>$50 & 25.64 & 50.0 & 17.24 \\
ORDINAL & $>$50 & 36.36 & 53.85 & 27.45 \\
ORG & $>$50 & 64.63 & 61.86 & 67.67 \\
PERCENT & 28 & 0.00 & 0.00 & 0.00 \\
PERSON & $>$50 & 59.88 & 66.60 & 54.38 \\
PRODUCT & $>$50& 4.68 & 20.00 & 2.65 \\
QUANTITY & $>$50 & 38.46 & 35.08 & 42.55 \\
TIME & $>$50 & 58.82 & 83.33 & 45.45 \\
WORK\_OF\_ART & $>$50 & 20.57 & 40.91 & 13.74 \\
\hline
\end{tabular}}
\caption{\label{naijaner-metrics}
Entity Evaluation Scores of the NaijaNER Model
}
\end{table}

\section{Discussion}
Models developed are available here\footnote{\url{https://git.io/JY0kk}} with an access to an interactive web app available here\footnote{\url{https://nigner.herokuapp.com/}}. 

Tables \ref{yorubaner-metrics}\---\ref{Ngenglishner-metrics} shows the metric scores observed for each of the NER models on the test data. There are certain entities in the entire dataset with low representation (less than 50), which resulted to the lower performance of the models when identifying such entities in the given test data. Therefore $>$50 was used for better performance.

The PidginNER and NigerianEnglishNER models were trained by leveraging the pretrained spaCy English model. This explains why both models performed better than the combined NaijaNER model on Nigerian Pidgin and English texts.

\begin{figure}[hbt!]
    \centering
    \includegraphics[width=8cm]{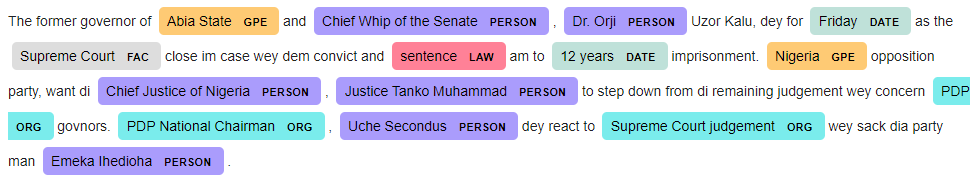}
    \caption{\label{pidginonpidgin}PidginNER on Nigerian Pidgin Text}
\end{figure}

\begin{figure}[hbt!]
    \centering
    \includegraphics[width=8cm]{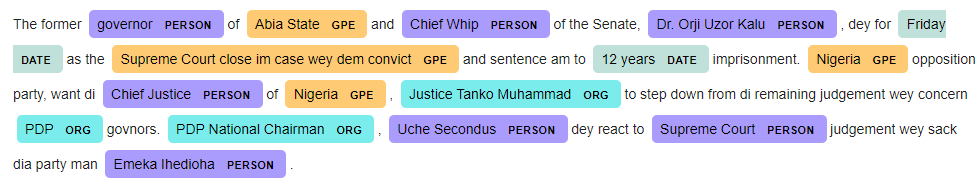}
    \caption{\label{naijaneronpidgin}NaijaNER on Nigerian Pidgin Text}
\end{figure}

For example, the PidginNER model \ref{pidginonpidgin} could identify every PERSON on the sample text, including "Justice Tanko Muhammad", which the NaijaNER wrongly identified as ORG. Also, the latter model \ref{naijaneronpidgin} misclassified "Supreme Court" as a PERSON as well as a GPE, but including other words. 

The figures displayed below compares the named entity recognition of the combined NaijaNER model to that of the individual models of respective native language sample texts. It can be stated that this combined model is robust enough to identify entites across various languages despite the small amount of indivdual langauge data this model was trained with.

\begin{figure}[hbt!]
    \centering
    \includegraphics[width=8cm]{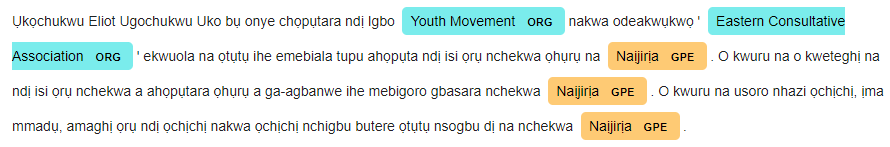}
    \caption{\label{igboonigbo}IgboNER on Igbo Text}
    
\end{figure}
\begin{figure}[hbt!]
    \centering
    \includegraphics[width=8cm]{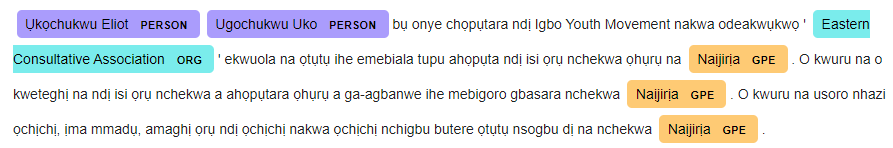}
    \caption{\label{naijaneronigbo}NaijaNER on Igbo Text}
\end{figure}

\begin{figure}[hbt!]
    \centering
    \includegraphics[width=8cm]{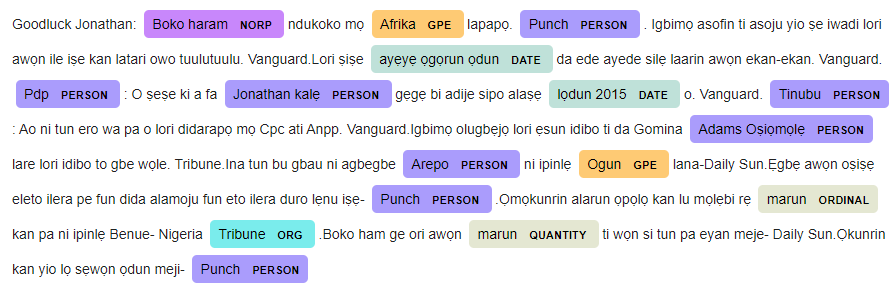}
    \caption{\label{yorubaonyoruba}YorubaNER on Yoruba Text}
\end{figure}

\begin{figure}[hbt!]
    \centering
    \includegraphics[width=8cm]{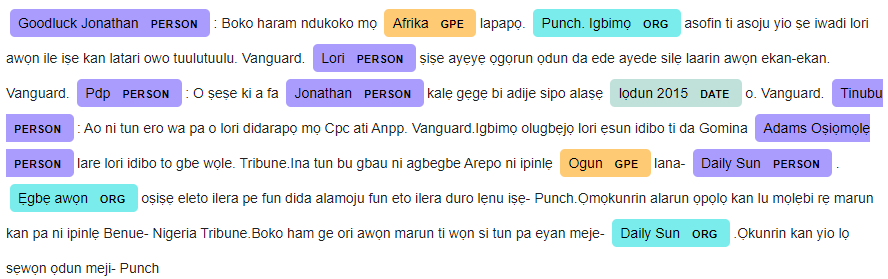}
    \caption{\label{naijaneronyoruba}NaijaNER on Yoruba Text}
\end{figure}

\begin{figure}[hbt!]
    \centering
    \includegraphics[width=8cm]{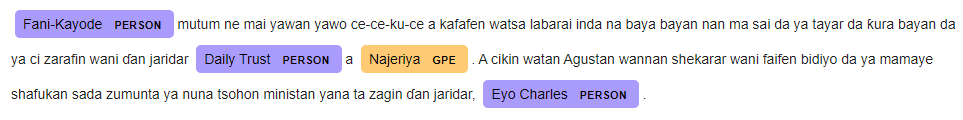}
    \caption{\label{hausaonhausa}HausaNER on Hausa Text}
\end{figure}

\begin{figure}[hbt!]
    \centering
    \includegraphics[width=8cm]{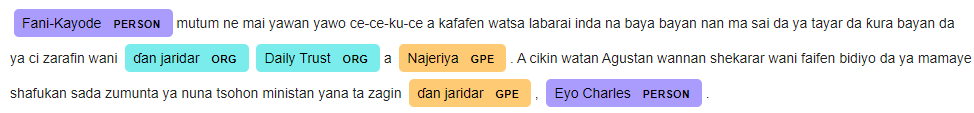}
    \caption{\label{naijaneronhausa}NaijaNER on Hausa Text}
\end{figure}

In the Igbo sample, the NaijaNER model \ref{naijaneronigbo} recognised Ukochukwu Eliot Ugochukwu Uko as PERSON, which was a better performance to the IgboNER model \ref{igboonigbo}. Also, for the Yoruba sample text, and just like the YorubaNER model \ref{yorubaonyoruba}, the NaijaNER model \ref{naijaneronyoruba} had some misindentifications of entities such as "Lori" (on), "Pdp" (a political party) and "Daily Sun" (a News company) as PERSON, whereas the rest of the entities were correctly named. Lastly, for the Hausa language, the NaijaNER model \ref{naijaneronhausa} was accurate in comparison to the HausaNER model \ref{hausaonhausa}, except for the unnecessary classification of "dan jaridar" (journalist) as ORG and GPE.

\begin{figure}[hbt!]
    \centering
    \includegraphics[width=8cm]{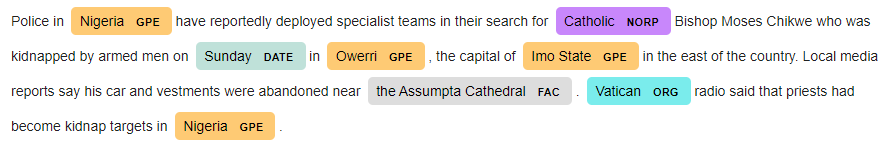}
    \caption{\label{ngneronngeng}NigeriaEnglishNER on English text in Nigerian context}
\end{figure}

\begin{figure}[hbt!]
    \centering
    \includegraphics[width=8cm]{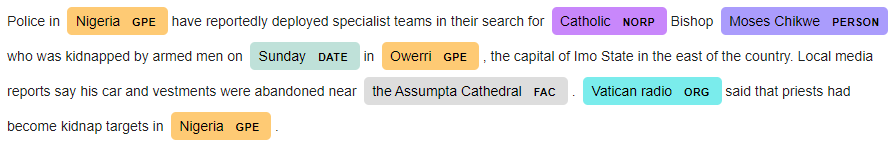}
    \caption{\label{spacyonngeng}spaCy English NER on English text in Nigerian context}
\end{figure}

These models are relevant to the context that Nigerian sentences have. This can be attributed to the purpose of developing the NigerianEnglishNER model. Figure \ref{ngneronngeng} shows that the NigerianEnglishNER model performed well in correctly identifying the entities in the text including "Imo State" (a State in Nigeria), that the spaCy en model \ref{spacyonngeng} could not recognise.

\section{Conclusion}
This work explores individual and combined Named Entity Recognition in 5 Nigerian languages - Nigerian English, Pidgin, Yoruba, Hausa and Igbo.


The choice of when to use a combined or individual models is dependent on the user's interest. The combined NER model-NaijaNER, is versatile for multi-language applications and has a good performance on each of the languages while individual language models showed better performance on their specific languages but will require deployment of multiple models.

Combined  Named  Entity  Recognition  that works with different languages can be a valuable tool in low resourced language systems and can help drive inclusion, ease of deployment in production and reusability of NER models.










\bibliography{anthology,eacl2021}
\bibliographystyle{acl_natbib}

\begin{center}

\end{center}

\end{document}